\documentclass[letterpaper, 10 pt, conference]{ieeeconf}
\IEEEoverridecommandlockouts 
\usepackage[backend=bibtex,
            hyperref=true,
            url=false,
            isbn=false,
            doi=false,
            backref=false,
            style=ieee,
            natbib=true,
            mincitenames=1,
            maxcitenames=1,
            citestyle=numeric-comp,
            sorting=nyt,
            block=none]{biblatex}
        
\usepackage{url}

\usepackage{hyperref}

\usepackage{amsmath,amssymb}
\usepackage{bm}

\numberwithin{equation}{section} 

\usepackage{graphicx}
\graphicspath{ {figures/} }
\usepackage{multicol,multirow}

\usepackage{tabularx}
\newcolumntype{Y}{>{\centering\arraybackslash}X}
\usepackage{booktabs}
\usepackage{caption}
\usepackage[table,dvipsnames]{xcolor}
\definecolor{green}{rgb}{0.34, 0.82, 0.45}
\definecolor{red}{rgb}{0.92, 0.05, 0.16}
\newcolumntype{b}{X}
\newcolumntype{s}{>{\hsize=.3\hsize}X}

\usepackage{mathtools}
\usepackage{autolabtools}

\title{\LARGE \bf
IPC-GraspSim: Reducing the Sim2Real Gap for Parallel-Jaw Grasping\\with the Incremental Potential Contact Model
}

\author{Chung Min Kim$^1$, Michael Danielczuk$^1$, Isabella Huang$^2$, Ken Goldberg$^1$%
\thanks{
$^{1}$The AUTOLab at University of California, Berkeley. \{chungmin99, mdanielczuk, goldberg\}@berkeley.edu
$^{2}$Department of Electrical Engineering and Computer Sciences, University of California, Berkeley, USA. isabella.huang@berkeley.edu
}}

\begin{document}

\maketitle


\begin{abstract}
Accurately simulating whether an object will be lifted securely or dropped during grasping is a longstanding Sim2Real challenge. Soft compliant jaw tips are almost universally used with parallel-jaw robot grippers due to their ability to increase contact area and friction between the jaws and the object to be manipulated. However, interactions between the compliant surfaces and rigid objects are notoriously difficult to model. We introduce IPC-GraspSim, a novel grasp simulator that extends Incremental Potential Contact (IPC) --- a highly accurate collision + deformation model developed in 2020 for computer graphics. IPC-GraspSim models both the dynamics and the deformation of compliant jaw tips
to reduce 
Sim2Real gap for robot grasping.
We evaluate IPC-GraspSim using a set of 2,000 physical grasps across 16 adversarial objects where analytic models perform poorly.
In comparison to both analytic quasistatic contact models (soft point contact, REACH, 6DFC) and dynamic grasp simulators (Isaac Gym with FleX), results suggest IPC-GraspSim can predict robustness 
with higher precision and recall (F1 = 0.85). IPC-GraspSim increases F1 score by 0.03 to 0.20 over analytic baselines and 0.09 over Isaac Gym, at a cost of 8000x and 1.5x more compute time, respectively.
All data, code, videos, and supplementary material are available at \url{https://sites.google.com/berkeley.edu/ipcgraspsim}.



\end{abstract}

\section{Introduction} \label{sec:introduction}
Recent work in grasping objects with parallel-jaw grippers uses deep neural networks to estimate high-quality grasps directly from RGB or depth images~\cite{zeng2018learning,pinto2016supersizing,kalashnikov2018scalable,mahler2019learning,lenz2015deep,morrison2020learning,fang2020graspnet}. These grasping networks are trained on large-scale datasets of labeled grasps generated either on physical systems~\cite{kalashnikov2018scalable,pinto2016supersizing,lenz2015deep,fang2020graspnet} or in simulation ~\cite{mahler2017learning,mahler2019learning,johns2016deep}. Simulation-based approaches can avoid time-intensive physical grasp attempts and have shown strong transfer to physical systems~\cite{mahler2019learning}, but may suffer from noisy labels due to errors in modeling the interaction between gripper jaws and objects.

Sim2Real for grasping is complicated due to 
soft material on the jaw tips
on almost all parallel-jaw robot grippers, from soft foam padding~\cite{xu2017grasping} to custom-made silicone rubber fingertips~\cite{mahler2019learning}.
Many simulators use discrete collision detection and penalty-based contact methods to determine contact forces at each timestep. Although computationally efficient, these simulators require that collision distances and timestep lengths be manually tuned for different object geometries and material parameters to avoid interpenetration or other physically unrealistic interactions between the jaws and object. Even so, object meshes may intersect during large timesteps or large contact force magnitudes, leading to false positive and false negative grasp predictions.

\begin{figure}
    \vspace{4pt}
    \centering
    \includegraphics[width=\linewidth]{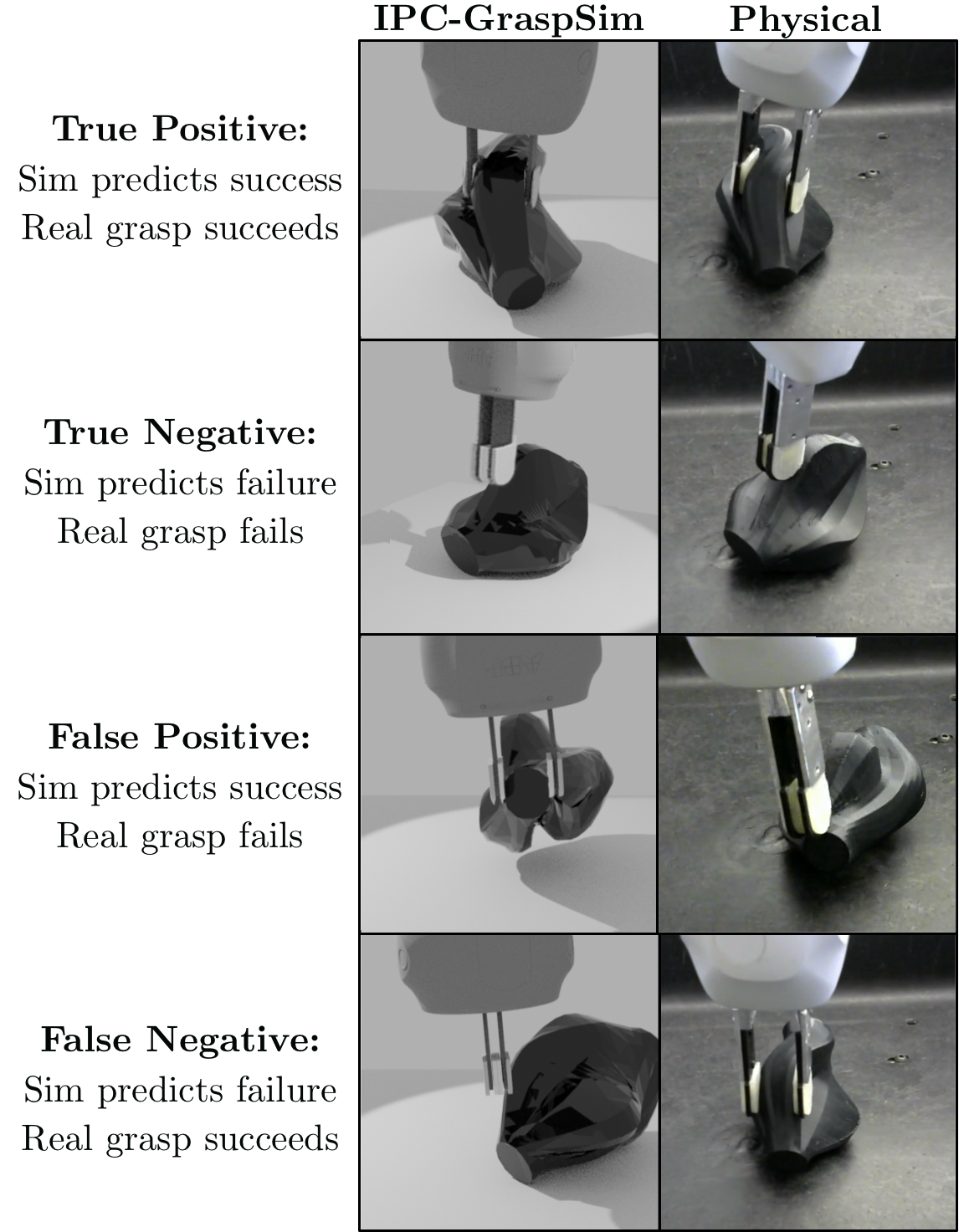}
    \caption{
    Simulated (left) and physical (right) grasp outcomes: in the first two rows, IPC-GraspSim correctly predicts grasp success and grasp failure. In the last two rows, predictions are incorrect due to unmodeled cantilever bending + rotations due to pushing. 
    }
    \vspace{-16pt}
    \label{fig:simreal}
\end{figure}

In 2020, \citet{li2020incremental} introduced Incremental Potential Contact (IPC), a simulator that uses a novel collision model and continuous collision detection method to dynamically step time and guarantee intersection-free simulation for elastodynamics contact problems with friction across variable material parameters and mesh resolutions. We adapt IPC to the problem of Sim2Real robot grasping of rigid objects with soft gripper pads to create IPC-GraspSim. IPC-GraspSim modifies IPC’s state machine by introducing a piecewise grasping action that 1) dynamically adapts the grasp width to the local object geometry by iteratively closing a set of gripper jaws until a deformation energy threshold is reached due to contact with the object, 2) lifts the jaws to a specified height, and 3) records grasp success based on the relative final positions of the jaws and object. IPC-GraspSim accurately models grasp outcomes produced by a physical system, as shown in Fig.~\ref{fig:simreal}. 

This paper makes three contributions:
\begin{enumerate}
    \item IPC-GraspSim: an application of the contact dynamics model introduced in Incremental Potential Contact~\cite{li2020incremental} to the Sim2Real robot grasping problem that models both friction and contact dynamics effects for compliant parallel-jaw gripper jaws. 
    \item An Isaac Gym-based parallel-jaw grasping simulator with FleX backend that extends~\citet{eppner2020acronym}, which only supports rigid gripper - rigid object interactions, by adding deformable gripper jaws.
    \item Experiments comparing IPC-GraspSim with Isaac Gym and 3 analytic contact models (soft point contact, REACH, and 6DFC), that suggest IPC-GraspSim can predict grasp robustness with F1 score of 0.85, an improvement in F1 score by 0.03 to 0.20 over analytic baselines and 0.09 over Isaac Gym, at a cost of 8000x more compute time.

\end{enumerate}

\section{Related Work} \label{sec:related work}
\subsection{Grasp Metrics and Contact Models for Soft Grippers}
Analytic grasp metrics use contact force and torque constraints generated by contact models to measure grasp robustness. As summarized by~\citet{roa2015grasp}, common metrics describe the ability of a grasp to resist external disturbances (force closure~\cite{bicchi1995closure,rimon1996force}), be at equilibrium between contact forces and external forces~\cite{kerr1986analysis}, or allow the object to resist gravity (wrench resistance~\cite{mahler2019learning,mahler2018dex}); these metrics can be determined using linear or quadratic programs based on the force and torque constraints at each contact.

Grasp contact models measure the ability of grasp contacts between the gripper jaw and the object to resist external forces and torques via the contact area, contact pressure and friction applied at the contact. Models of compliant gripper contacts typically use the quasistatic approximation. \citet{bicchi2000robotic} summarize analytic physics-based contact models, including soft point contact models \cite{salisbury1983kinematic,ciocarlie2005grasp,howe1988sliding,kao1992quasistatic,kao2004stiffness}, which exert forces in the plane tangent to the contact surface and a torsional moment about the contact normal. Frictional limit surface models jointly limit the tangential force and torque about the contact normal~\cite{goyal1991planar,li2001review,howe1988sliding,zhou2018convex} and can be extended to model non-planar contacts~\cite{xu20216dls}. Another approach is to approximate contact area with geometric primitives, such as planes, spheres, or cylinders, that have analytic solutions~\cite{hertz1896miscellaneous,ciocarlie2007soft}. Similarly, \citet{danielczuk2019reach} and \citet{xu20206dfc} discretize the contact area into a triangular mesh and formulate per-triangle or per-contact limit surface constraints based on a static friction assumption. In contrast to these quasistatic grasp contact models and metrics, we use a simulator that incorporates both grasp dynamics and frictional area contacts.

\subsection{Grasp Simulators for Soft Grippers}
Grasp simulators, unlike grasp contact models, can incorporate dynamics which are crucial to grasping: they can predict which jaw makes contact first, and how that affects post-grasp displacement as the gripper and object settle into a stable position~\cite{zhao2020towards}. 
In this paper, we focus on dynamic simulators that model deformable materials~\cite{miller2004graspit,leon2010opengrasp,coumans2021,liang2018gpu}, as it is important to model frictional contact dynamics simultaneously with gripper compliance: deformation affects surface area, which in turn affects object dynamics. 

Isaac Gym is a GPU-enabled simulator favored by the robotics community for its speed and python-based simulator with a straightforward robotics interface. NVIDIA FleX engine integration has recently made deformables available for robotics environments, as evidenced by recent papers that model the SynTouch BioTac~\cite{narang2021sim} and grasping of deformable objects with rigid grippers~\cite{huang2021defgraspsim}. However, many simulators, including Isaac Gym, require manual parameter tuning to avoid interpenetration artifacts and may be sensitive to friction and closing force parameters~\cite{rocchi2016stable}.

In simulation, grasp success may be defined as (1) the object can be lifted above the surface~\cite{mahler2017learning, huang2021defgraspsim}, (2) the object can be lifted and withstand perturbations via shaking motions~\cite{eppner2019billion,eppner2020acronym}, (3) the object is removed from its environment after grasp and move actions~\cite{zeng2018learning}, or (4) the object's post-grasp pose matches a desired pose~\cite{allshire2021transferring}. In this paper, we use the first definition of grasp success.

\subsection{Simulation of Deformable Materials} \label{sec:def_sim}
Simulating deformable materials is challenging due to their many degrees of freedom and friction between deformable and rigid objects. However, due to the high computational load in more accurate representations such as Finite Element Methods (FEM), most work relies on simplified models as provided by mass-spring systems and position-based dynamics, which are computationally efficient but do not have a rigorous mathematical model for contact and friction~\cite{bender2014survey}. With FEM (e.g., PyBullet's FEM option and graphics implementations), Neo-Hookean materials are often used to simulate hyperelastic materials where the stress-strain relationships are highly nonlinear, such as silicone rubber~\cite{marckmann2006comparison}. Isaac Gym models softbody dynamics using corotational linear-elastic constitutive models, and leverages a custom GPU-based solver~\cite{macklin2019non}.


A common model for applying loading conditions onto the deformable bodies is the penalty-based contact, which calculates contact force as proportional to the mesh penetration depth and penalty value $p$, assuming nonzero interpenetration for all contacting bodies. High values of $p$ can decrease the penetration depths, but also make the problem ill-conditioned~\cite{04ccf53210f44e1c983e90e4c02e7816}. Furthermore, any contact violation can cause catastrophic failures~\cite{li2020incremental}. A common way to handle interpenetration is to add a boundary layer on the deformable to ``thicken" the mesh for collisions, but the layer depth must be manually tuned for each scenario~\cite{liang2018gpu,rocchi2016stable}. 

Incremental Potential Contact (IPC) \cite{li2020incremental} is a computer graphics model and algorithm presented in 2020 based on a custom nonlinear Newton-based solver that solves elastodynamics contact problems with the smoothed barrier method. It guarantees a collision-free state at every timestep to provide robust solutions for all choices of material parameters, timestep sizes, impact velocities, deformation severities, and enforced boundary conditions. IPC provides inversion-free and intersection-free solutions, so that objects do not penetrate and remain stuck to one another in future timesteps. IPC accounts for static friction with a smoothed approximation to eliminate an explicit Coulomb constraint, which leads to the assumption that static and dynamic friction coefficients are equal between each material combination. The authors justify this decision by implementing an approximation to static friction that can successfully model stiction.

We leverage IPC as the physics engine of a grasping simulator that avoids the interpenetration issues described in Section \ref{sec:def_sim} and more accurately models grasp outcomes observed on a physical system than existing quasistatic or dynamic simulator approaches.

\section{Problem Statement} \label{sec:problem statement}
We evaluate the ability of IPC-GraspSim to accurately predict the grasp outcomes in a physical grasp dataset with known object and gripper geometries and known grasp poses.

\subsection{Assumptions}
We make the following assumptions:
\begin{enumerate}
    \item A singulated rigid object of known geometry and uniform density rests on a planar surface. 
    \item The gripper has known geometry and two parallel jaws.
    \item The silicone rubber material on the gripper jaws can be simulated with the Neo-Hookean material model.
\end{enumerate}

\subsection{Definitions}
The true state \textbf{x} of the grasping environment includes geometry (including deformations), pose, mass, and frictional properties of the object and gripper jaws.
Each grasp \textbf{u} is represented by the parallel gripper jaw center and the grasp axis orientation $\varphi \in \mathbb{R}^3$. 
The modifiable gripper jaw parameters include the deformable material elasticity and the friction coefficient between the gripper jaw pads and the grasping object.
Success $S(\mathbf{x}, \mathbf{u})$ is measured with a binary reward function $S$, where $S=1$ if the grasp successfully lifts the object and $S=0$ otherwise.

To model uncertainty and imprecision in robot control, we define grasp robustness $R(\mathbf{x}, \mathbf{u})=\mathbb{E}[S(\mathbf{x}, \mathbf{u'})]$, where the perturbed grasp pose $\mathbf{u'} = \mathbf{u} + \varepsilon$ is offset by $\varepsilon$ sampled from a normally-distributed pose random variable. To calculate robustness in our experiments, we use Monte-Carlo sampling to estimate $R$ with the sample mean of $N$ trials each with a perturbed gripper pose: $R=\frac{1}{N} \sum_{i=1}^{N} S(\mathbf{x}, \mathbf{u'}_{i})$. We emphasize the importance of using sampling with grasp trials; although the robot may intend the same grasp on a given object, not all trials will behave equally due to randomness introduced by pose estimation, robot control, and frictional forces. 

\subsection{Metrics}
To quantify how closely the predicted grasp robustness from the simulator matches the robustness measured on the physical system, we measure the F1 score, a harmonic mean of average precision (AP) and average recall (AR)
. F1 score is a common measure of binary classification performance for imbalanced datasets in computer vision~\cite{saito2015precision}.

\begin{figure}
    \vspace{4pt}
    \centering
    \includegraphics[width=\linewidth]{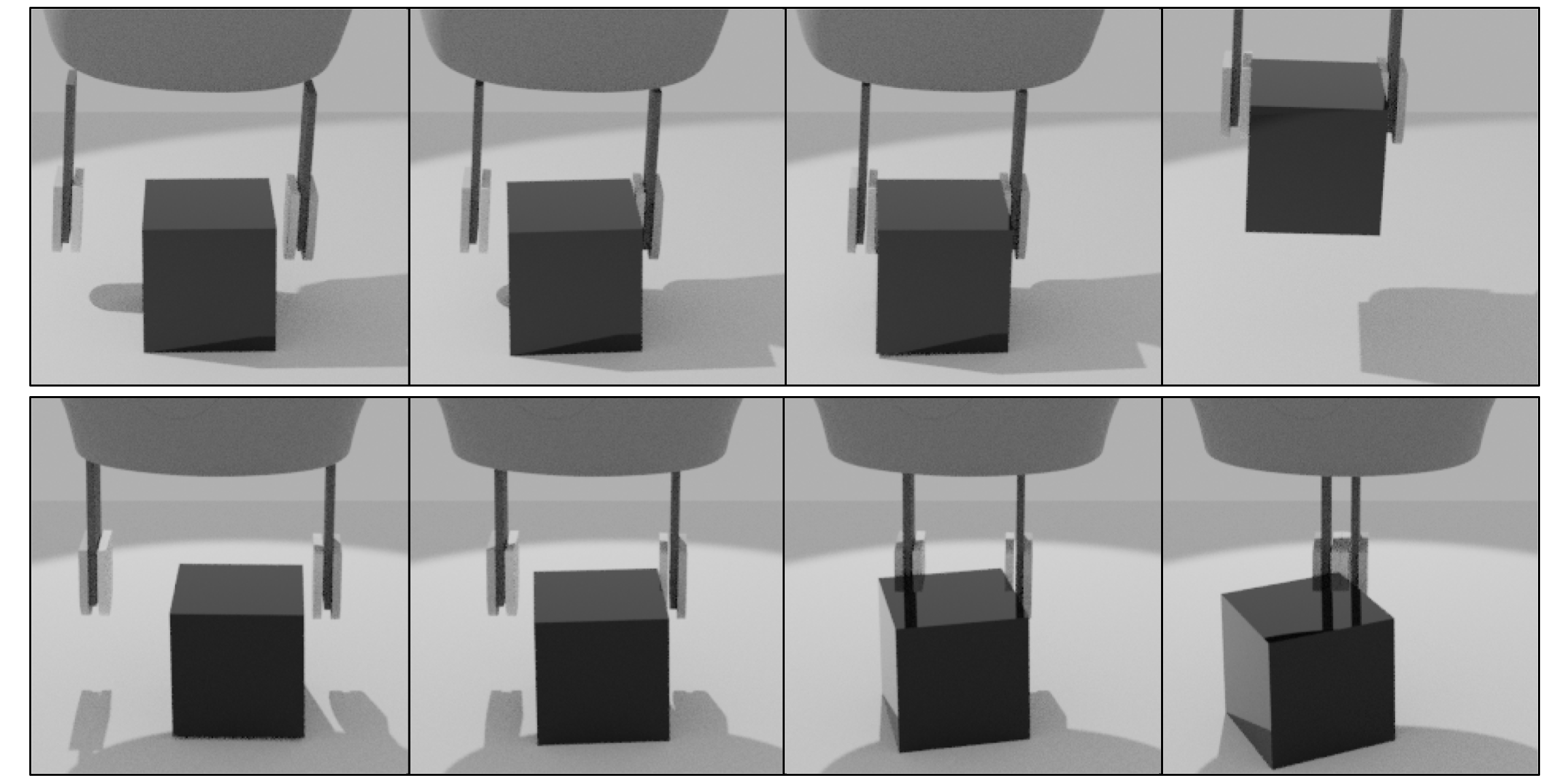}
    \caption{
    Two grasps on a cube with IPC-GraspSim, showing state at successive timesteps. (Top row) Successful grasp: 1) cube and gripper initialized; 2) cube contacts right jaw; 3) cube contacts left jaw; 4) gripper lifts cube.
    (Bottom row) Failed grasp: 1) cube and gripper initialized; 2) cube contacts right jaw; 3) cube \textit{misses} left jaw; 4) jaws meet each other.
    }
    \vspace{-10pt}
    \label{fig:prog_cube}
\end{figure}


\section{IPC-GraspSim} \label{sec:ipc-graspsim}
To simulate grasps, we apply Incremental Potential Contact \cite{li2020incremental}, as it guarantees intersection-free and inversion-free solutions for all parameter choices, as described in Section~\ref{sec:def_sim}. This model avoids the interpenetration artifacts faced by many simulator models~\cite{coumans2021, liang2018gpu}, which include false sticking behavior after gripper lifting (false positive) or intersecting through the object (false negative). To avoid these large interpenetration errors in other simulators, collision parameters need to be carefully tuned. 

As IPC is originally created for computer graphics applications, IPC-GraspSim implements a robotic grasp interface: the inputs for IPC-GraspSim are the grasping state $\mathbf{x}$ and grasp $\mathbf{u}$. Then the simulator performs a perturbed grasp $\mathbf{u'} = \mathbf{u} + \varepsilon$ where $\varepsilon$ is noise in grasp pose, with translation noise (in meters) sampled from $\mathcal{N}(0, 0.001)$ and rotation noise (in radians) from $\mathcal{N}(0, 0.003)$. IPC-GraspSim returns grasp evaluation $S(\mathbf{x}, \mathbf{u'})$ upon the end of the simulation, and we retrieve robustness $R(\mathbf{x}, \mathbf{u})$ over multiple simulation trials with different perturbations. We also implement a finite state machine in the original IPC simulator in order to progress through a piece-wise grasping motion that actively adapts to the grasp state $\mathbf{x}$, which is detailed in the following section. No changes are made to IPC's  contact dynamics model.

\subsection{Grasp Simulation Procedure}
IPC-GraspSim models grasping with a velocity-controlled compliant parallel-jaw gripper as follows:
\subsubsection{Initialize}
Initialize gripper jaws and object location as specified by the perturbed grasp $\mathbf{u'}$ and state \textbf{x}. Gripper jaws are opened to their max grasp width, and unobstructed by any object other than the target. The gripper is modeled as two identical jaws, each composed of an extruded compliant polygon pad attached to a rigid rectangular plate. Figures \ref{fig:simreal} and \ref{fig:prog_cube} show the compliant padding in light gray, and rigid backing in dark gray. 

\subsubsection{Squeeze} 
Close gripper pads along the grasp axis with a constant velocity. The target object will interact with the gripper pads and the ground surface, and is free to translate or rotate. 
Although YuMi, the physical robot used in this paper, completes the gripper closing action when it reaches a maximum specified closing force, contact force information is currently unavailable in the IPC simulator. To account for this, IPC-GraspSim approximates closing force on the target object by checking if A) the deformation energy $\Psi$ described in \citet{li2020incremental} in both gripper tooltips $\Psi_1,\Psi_2$ have increased such that $\Psi_1,\Psi_2\geq\Psi_{th}$, and B) the energy is similar in both gripper jaws $\Psi_1\approx\Psi_2$. $\Psi_{th}$ is a user-defined threshold to limit the amount of deformation in each pad $(\Psi_1,\Psi_2)$ and corresponds to the $\Psi$ that best matches maximum gripper closing force for the specified gripper. (A) assures that the closing action will not incorrectly terminate due to deformations driven by other external forces (e.g., gravitational), and (B) assures that equal effort is applied by both grippers. We test various values of $\Psi_{th}$ on a range of grasps. Based on visual comparison with our physical silicone gripper pads, we choose $\Psi_{th}=5\times 10^{4}$. Further details on $\Psi_{th}$, including sensitivity analysis, are in the supplement. 



\subsubsection{Lift} \label{sec:grasp-sim-lift}
Once the deformation energy threshold is reached, the two gripper jaws are lifted upwards with the jaw distance held constant until the object's lowest point is at least 2 cm above the plane by the time the simulation finishes.
A simulated grasp is considered successful if both gripper jaws are in contact with the object at the end of simulation, and the object is not in contact with the plane. Examples of successful and failed grasps are shown in Figure \ref{fig:prog_cube}.

\begin{table}
    \vspace{4pt}
    \centering
    \begin{tabularx}{\linewidth}{clYYYY} \toprule
    \multirow{7}{1.2cm}{\centering \includegraphics[width=\linewidth]{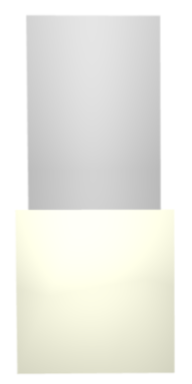}} & 
    \multirow{2}{2cm}{\centering Young's modulus $E$ (Pa)} & 
    \multicolumn{4}{c@{}}{\text{Friction coefficient $\mu$}} \\
    \cmidrule(l){3-6} & & 0.3 & \textbf{0.4} & 0.5 & 0.6 \\ \cmidrule(l){2-6}
    & 1e7            & \cellcolor{red!10}0.77 & \cellcolor{green!20}0.82	& \cellcolor{green!30}0.83 & \cellcolor{green!30}0.84 \\
    & \textbf{1e8}   & \cellcolor{green!40}0.85 & \cellcolor{green!50}\textbf{0.86} & \cellcolor{green!50}0.80 & \cellcolor{green!10}0.78 \\
    & 1e9            & \cellcolor{green!40}0.84 & \cellcolor{green!20}0.81	& \cellcolor{red!10}0.77 & \cellcolor{red!40} 0.74 \\
    & 1e10  & \cellcolor{red!10}0.77 & \cellcolor{green!10}0.78	& \cellcolor{red!30} 0.75 & \cellcolor{red!40} 0.74 \\
    & 1e11  & \cellcolor{red!10}0.77 & \cellcolor{red!40}0.74 & \cellcolor{red!50} 0.73 & \cellcolor{red!40} 0.74 \\ \bottomrule
    
    \end{tabularx}
    \caption{\textbf{Rectangular} jaw: results of parameter grid search over 3 training objects. $(E, \mu)=(10^8, 0.4)$ has the highest F1 score of 0.86.
    }
    \label{tab:sharp_jaw_gridsearch}
    \smallskip
    \centering
    \begin{tabularx}{\linewidth}{clYYYY} \toprule
    \multirow{7}{1.2cm}{\includegraphics[width=\linewidth]{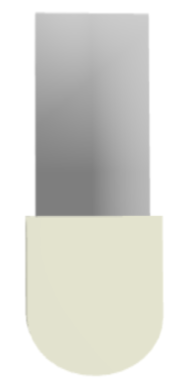}} & 
    \multirow{2}{2cm}{\centering Young's modulus $E$ (Pa)} & 
    \multicolumn{4}{c@{}}{\text{Friction coefficient $\mu$}} \\
    \cmidrule(l){3-6} & & 0.3 & \textbf{0.4} & 0.5 & 0.6 \\ \cmidrule(l){2-6}
    & 1e7  & \cellcolor{red!50}0.66	& \cellcolor{red!40}0.70	& \cellcolor{red!40}0.70	& \cellcolor{red!30}0.74 \\ 
    & \textbf{1e8}  & \cellcolor{green!20}0.85	& \cellcolor{green!50}\textbf{0.91}	& \cellcolor{green!40}0.89	& \cellcolor{green!15}0.85 \\
    & 1e9  & \cellcolor{green!20}0.85	& \cellcolor{red!10}0.82	& \cellcolor{green!30}0.86	& \cellcolor{green!10}0.83 \\ 
    & 1e10 & \cellcolor{red!20}0.82	& \cellcolor{green!40}0.88	& \cellcolor{green!30}0.87	& \cellcolor{green!10}0.84 \\
    & 1e11 & \cellcolor{red!20}0.82	& \cellcolor{green!10}0.84	& \cellcolor{red!10}0.81	& \cellcolor{green!30}0.86 \\ \bottomrule
    \end{tabularx}
    \caption{\textbf{Rounded} jaw: results of parameter grid search over 3 training objects. $(E, \mu)=(10^8, 0.4)$ has the highest F1 score of 0.91. 
    }
    \vspace{-12pt}
    \label{tab:round_jaw_gridsearch}
\end{table}

\section{IPC-GraspSim: Parameter Sensitivity Analysis}
\label{sec:IPC-Sensitivity}
We analyze the sensitivity of IPC-GraspSim to the elasticity~$E$ and dynamic friction~$\mu$ of the gripper pad material using a coarse grid search on the parameter space. We also test two grippers with different geometries, as shown in Table~\ref{tab:sharp_jaw_gridsearch} (rectangular) and Table~\ref{tab:round_jaw_gridsearch} (rounded), to study the effect of accurately modeling the gripper geometry.

\subsection{Test Object Set} \label{sec:test_objs}
In our experiments, we use the physical dataset from \citet{danielczuk2019reach}: a set of 2,000 grasps across a set of 16 3D-printed adversarial objects on a physical ABB YuMi robot with a compliant parallel-jaw gripper~\cite{guo2017design}. 

We evaluate grasp outcomes on 3 objects with 1) a similar number of successful and unsuccessful grasps and 2) varied geometries. Then, we identify the set of parameters $(\hat{E}, \hat{\mu})$ with the highest F1 score in the training set, and use these parameters to simulate grasps on the remaining 13 objects in Section \ref{sec:Experiments}. 

\subsection{Sensitivity to Young's Modulus and Friction}
We vary the Young's modulus of the jaw pad from $10^{7}$ Pa (rubbery material) to $10^{11}$ Pa (metal material). We linearly vary the friction coefficient between the pad and object from 0.3 to 0.6. Note that the exact friction between the jaw pad and object is unknown due to unknown material properties and the unmodeled surface variations on the jaw pad, as shown in Dex-Net's silicone rubber fingers in Figure~\ref{fig:comp_gripper_wild}. The millistructure design results in increased friction between the jaw pad and the object~\cite{guo2017design}. We aim to find both the Young's modulus and single friction coefficient that closely models such a structure.

For both gripper jaw models, lower Young's modulus values $(E, \mu)=(10^8,0.4)$ have higher AP, AR, and F1 score as compared to higher Young's modulus values $(10^{11},0.4)$. For rounded jaws, more compliant jaws increase AP by 13\%, AR by 2\%, and F1 by 7\%; for rectangular jaws, they increase AP by 14\%, AR by 11\%, and F1 by 12\%. 
Increase in $\mu$ appears to correlate with an increase in AR, as the additional frictional forces allow for a greater number of successful grasps; however, AP drops as the number of false positives increases. 
%
We hypothesize the increase in AR occurs due to the increased contact area in compliant grippers, allowing the grippers to deform around the contact point and distribute contact forces across the object's surface. However, when $\mu$ is too high, objects no longer can slip out of the gripper jaws and the number of false positive predictions increases. 

\begin{table*}
    \centering
    \vspace{4pt} 
    \includegraphics[width=\textwidth]{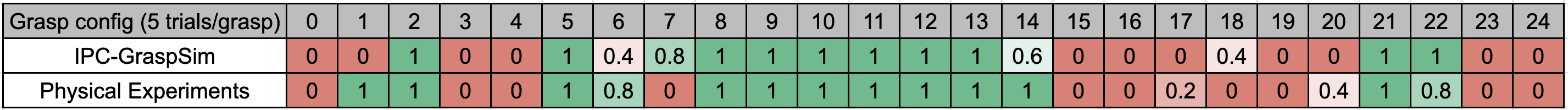}
    \caption{
    Grasp robustness values predicted by IPC-GraspSim (top row) and observed from 125 physical experiments (bottom row) using an ABB YuMi with a compliant parallel-jaw gripper on the ``vase" object shown in Fig.~\ref{fig:simreal}. 
    Each column corresponds to a unique grasp configuration (i.e., relative pose between gripper and object).
    Each cell contains the robustness $R=\frac{1}{N} \sum_{i \in \{1..N\}}S_i$, with $N=5$, and is colored to indicate its relative robustness (red: unreliable, green: robust).
    We observe a very high correlation of predicted grasp robustness between the two rows. Example true positive (grasp 2), true negative (grasp 0), false positive (grasp 7), and false negative (grasp 1) grasps are shown in Fig.~\ref{fig:simreal}. 
    See Section~\ref{sec:fail} for discussion of grasp configurations 1 and 7.
    }
    \label{fig:vase-sim-real-correlation}
    \vspace{-8pt}
\end{table*}

\subsection{Gripper Geometry}
The compliant rounded jaws result in a 17\% increase in AP and 8\% decrease in AR, leading to an overall increase of 5\% in F1 score, compared to the compliant rectangular jaws when using the highest-performing parameters for each geometry $(E = 10^8, \mu = 0.4)$. We hypothesize that the slight precision-recall tradeoff occurs as the smaller cross-sectional area of the rounded gripper jaw causes grasp robustness predictions to be more conservative. However, the F1 score ultimately increases, as the extra contact area due to the rectangular jaws had allowed grasps to incorrectly succeed.

Based on the parameter sensitivity analysis, we use the rounded gripper jaw model with $(\hat{E},\hat{\mu})=(10^8, 0.4)$ to model the compliant gripper jaws in dynamic simulation. We note that this optimal set of parameters closely matches the physically estimated parameters ($10^7-10^8$ Pa, 0.46). 

\section{Experiments} \label{sec:Experiments}

\begin{figure*}
    \centering
    \vspace{4pt}
    \includegraphics[width=\linewidth]{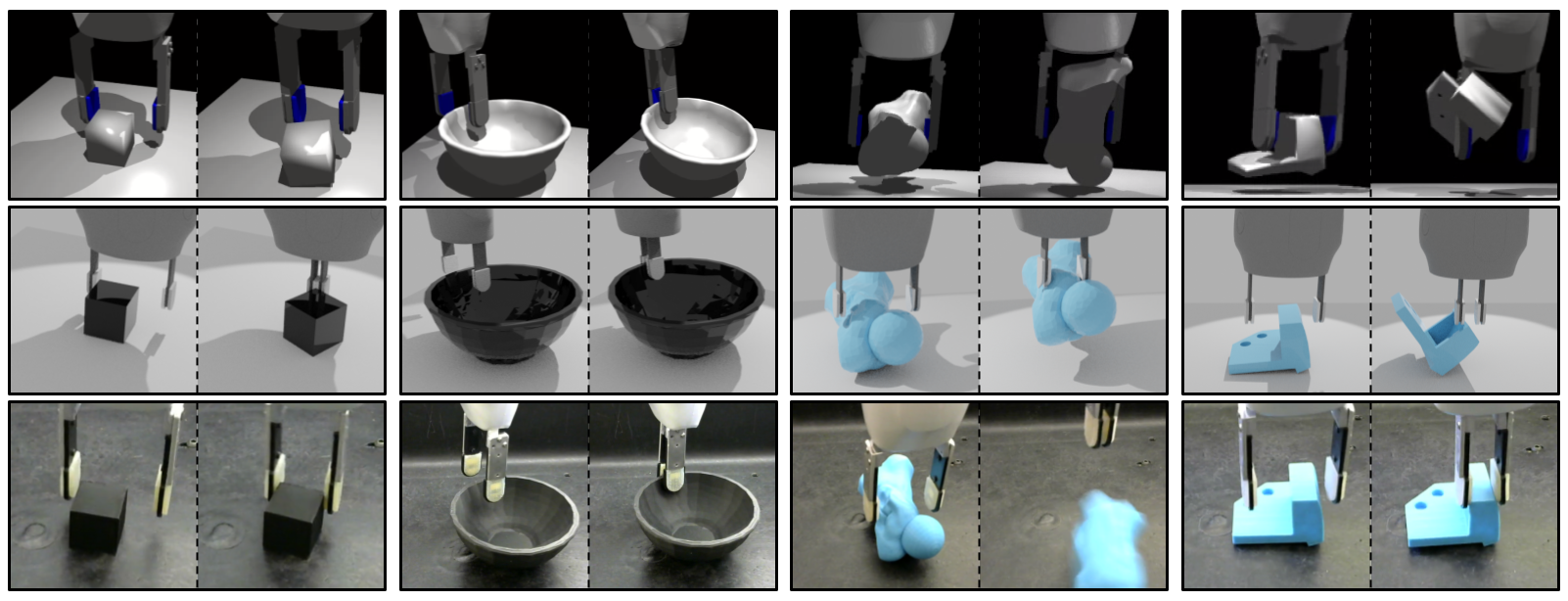}
    \caption{
    Four grasps (columns) with three simulators: (Top to bottom) Isaac Gym, IPC-GraspSim, Physical Experiments. 
    (A) False negative: IPC-GraspSim and Isaac Gym fail to predict that the right jaw will contact the cube.
    (B) True positive: bowl correctly moves and rotates into gripper for both sims.
    (C) False positive: In Isaac Gym, incorrect contact forces push ``pawn" up and away from the plane. In IPC-GraspSim, rigid jaws have sufficient contact area to grasp the object. Physical grasp fails as jaw cantilevering decreases contact area. 
    (D) False negative: For Isaac Gym, the object is pushed upwards, with a similar issue as described in (C). In IPC-GraspSim, left gripper jaw pushes object out of the gripper. 
    Videos of all simulator grasp attempts can be found on the website.
    }
    \vspace{-10pt}
    \label{fig:examples}
\end{figure*}

\subsection{Experiment Setup}
To evaluate the precision and recall of IPC-GraspSim, we use the 13 held-out objects from physical dataset from~\citet{danielczuk2019reach}. The objects in this dataset were chosen because they span a wide range of object geometries and grasp success rates. All objects are 3D printed such that simulated and physical attempted grasps are identical.
For benchmark experiments, we run 5 trials for each grasp while perturbing the grasp pose in each trial for dynamics randomization. The trials are then averaged to calculate the grasp robustness predicted by each grasp model.

\subsection{Benchmark Grasp Models}
In addition to IPC-GraspSim, we benchmark 4 other models; three analytic models used in the literature (soft point contact model, REACH~\cite{danielczuk2019reach}, 6DFC~\cite{xu20206dfc}), and a dynamic softbody simulation implemented in NVIDIA Isaac Gym with Flex backend~\cite{liang2018gpu}. We 
measure the performance of Isaac Gym and IPC-GraspSim by simulating grasps with the rounded gripper jaws and material properties $(E=10^8, \mu=0.4)$ determined in Section~\ref{sec:IPC-Sensitivity}. Soft Point, REACH, 6DFC, and IPC-GraspSim are run on an Ubuntu 18.04 on a single core of a Dual 20-Core Intel Xeon E5-2698 v4 2.2 GHz, and Isaac Gym on an Ubuntu 20.04 machine with an Intel i7-6850K processor and NVIDIA Titan X GPU.

\begin{table}
    \centering
    \begin{tabularx}{\linewidth}{bsssc} \toprule
       Model & AP    & AR    & F1    & Runtime (s) \\ \midrule
        
        Soft Point & 0.89  & 0.59  & 0.71  & \textbf{0.02}  \\
        REACH~\cite{danielczuk2019reach} & 0.75  & 0.57  & 0.65  & 0.44  \\
        6DFC~\cite{xu20206dfc} & 0.82  & 0.83  & 0.82  & 0.07 \\
        
        Isaac Gym     & 0.75 &	0.79	& 0.76  & 421            \\
        IPC-GraspSim    & \textbf{0.86}  & \textbf{0.84}  & \textbf{0.85}  & 605            \\ \bottomrule
    \end{tabularx}
    \caption{Average Precision (AP) and Average Recall (AR) for each model's grasp quality predictions for the 2000 grasps collected on the physical robot. All runtimes are provided per-grasp. Per-frame runtimes for Isaac Gym and IPC-GraspSim are 0.28 and 6.05 seconds respectively.
    }
    \vspace{-10pt}
    \label{tab:dset_results}
\end{table}

\subsection{Results}
The results in Table~\ref{tab:dset_results} 
suggest IPC-GraspSim can predict grasp robustness with F1 score of 0.85, an improvement in F1 score by 0.03 to 0.20 over analytic baselines and 0.09 over Isaac Gym, at a cost of 8000x more compute time. We observe an average increase of 0.17 in average recall in IPC-GraspSim compared to grasp models that approximate compliance (Soft Point, REACH, 6DFC) and a 0.05 increase in average recall compared to Isaac Gym. 

We hypothesize that the increase in recall as compared to the analytic models is due to the ability of IPC-GraspSim to model dynamics and compliance simultaneously. Modeling dynamics results in correct predictions for grasps that rotate into alignment with the gripper, a common failure mode for quasistatic models~\cite{danielczuk2019reach,xu20206dfc}. In addition, modeling compliance keeps the object closer to the gripper as it moves by 1) conforming to the shape of the object and 2) applying normal and frictional forces that stabilize the object. The combination of dynamics and compliance models the changing contact area and contact pressure over the course of the grasping motion. 

We also note that despite modeling dynamics, Isaac Gym has F1 score 0.06 lower than 6DFC, a quasistatic contact model; this result suggests that if the underlying simulator for modeling compliant gripper grasps is not sufficiently robust (i.e., simulator returns sudden large forces due to incorrect collision force handling), it may perform worse than a robust model that ignores dynamics. Although IPC-GraspSim only increases F1 score by 0.03 compared to 6DFC~\cite{xu20206dfc}, the simulator has significant room for improvement: most of its failure modes are due to unmodeled physical effects (e.g., gripper jaw cantilever effects, further detailed in Section~\ref{sec:fail}), which we hope to address in future work.

In comparison to Isaac Gym, IPC-GraspSim increases average precision by 0.11, average recall by 0.05, and F1 score by 0.09. The performance of IPC-GraspSim highlights the importance of intersection-free and robust modeling of soft contacts for compliant gripper modeling. First, poorly-resolved contacts may cause grasped objects to move unpredictably (e.g., float or swing into air, as shown in Figure~\ref{fig:examples}C-D for Isaac Gym). Second, incorrect contacts introduce force imbalance between the two compliant jaws and cause the target object to wobble around or become lodged in the gripper, resulting in unrealistic object behavior.

Although dynamic simulators (Isaac Gym, IPC-GraspSim) run at least three orders of magnitudes slower than analytic grasp models, these evaluations are intended to run offline for building large-scale datasets. Therefore, these simulations can be performed in parallel over extended periods of time.



\subsection{Failure Mode Analysis} \label{sec:fail}
Table~\ref{fig:vase-sim-real-correlation} shows results for each grasp attempt of the ``vase" (object 21 in the grasp dataset collected by \citet{danielczuk2019reach}, shown in Figure~\ref{fig:simreal}). A full breakdown of grasp attempts and predictions for all objects can be found in the supplement. For this object, we observe a high correlation between IPC-GraspSim predictions and physical experiments. IPC-GraspSim achieves an F1 score of 0.85 as compared to 0.65-0.82 over analytic methods and 0.68 for Isaac Gym.

We observe a false positive prediction for grasp 7, where gripper jaws bend and cantilever away from the object under the stress from the closing forces. The contact forces then push the vase away from and out of the gripper. However, in simulation, the gripper backings are perfectly rigid, and and do not account this bending phenomenon.

We also observe a false negative prediction for grasp 1, where the right jaw contacts the vase first and rotates the vase out of the gripper. This phenomenon is also observed in Figure~\ref{fig:examples}A. This missed prediction suggests that precise gripper geometry and gripper closing speed have an impact on grasp predictions. We additionally hypothesize that the dynamics between the ground plane and the object are not well-modeled in simulation.
\section{Discussion}
In this paper, we present IPC-GraspSim, an intersection-free, robust grasp simulator, and evaluate its performance on a physical grasp dataset generated with compliant parallel-jaw grippers. Compared with analytic baselines, IPC-GraspSim improves recall while maintaining similar precision, which we hypothesize is a result of modeling the compliance in a dynamic context: the contact points between the grippers and the target object change throughout the entire gripper closing process. IPC-GraspSim also improves precision over Isaac Gym due to its ability to stably model soft contacts without interpenetration. IPC is designed for simulating contact dynamics, and thus puts speed as lower priority. In future work, we will accelerate the simulator by using Medial-IPC~\cite{lan2021medial}, which has demonstrated speed performance improvements up to 110x, and use it to model gripper jaw cantilever + object pushing effects.
We will explore the ability of IPC-GraspSim to generalize to different jaw materials and geometries.
We will also measure the simulator's performance for grasps with different gripper and object geometries, and explore the effect of training grasp prediction neural networks using IPC-GraspSim.



\section{Appendix}

\subsection{Stochasticity in Sim and Real} \label{sec:stochastic}
\begin{figure}
    \centering
    \includegraphics[width=0.5\linewidth]{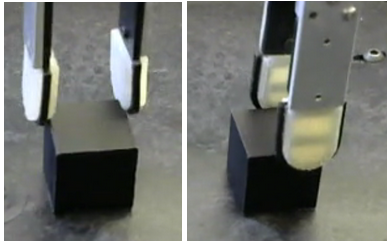}
    \caption{Corner grasp on cube: (left) gripper perturbed away from object, physical grasp attempt fails (right) gripper perturbed towards the object, physical grasp attempt succeeds. Both grasp trials intend to perform the same action, yet uncertainties allow the grasp success to vary across trials.}
    \label{fig:corner_grasp}
    \smallskip
    \includegraphics[width=0.8\linewidth]{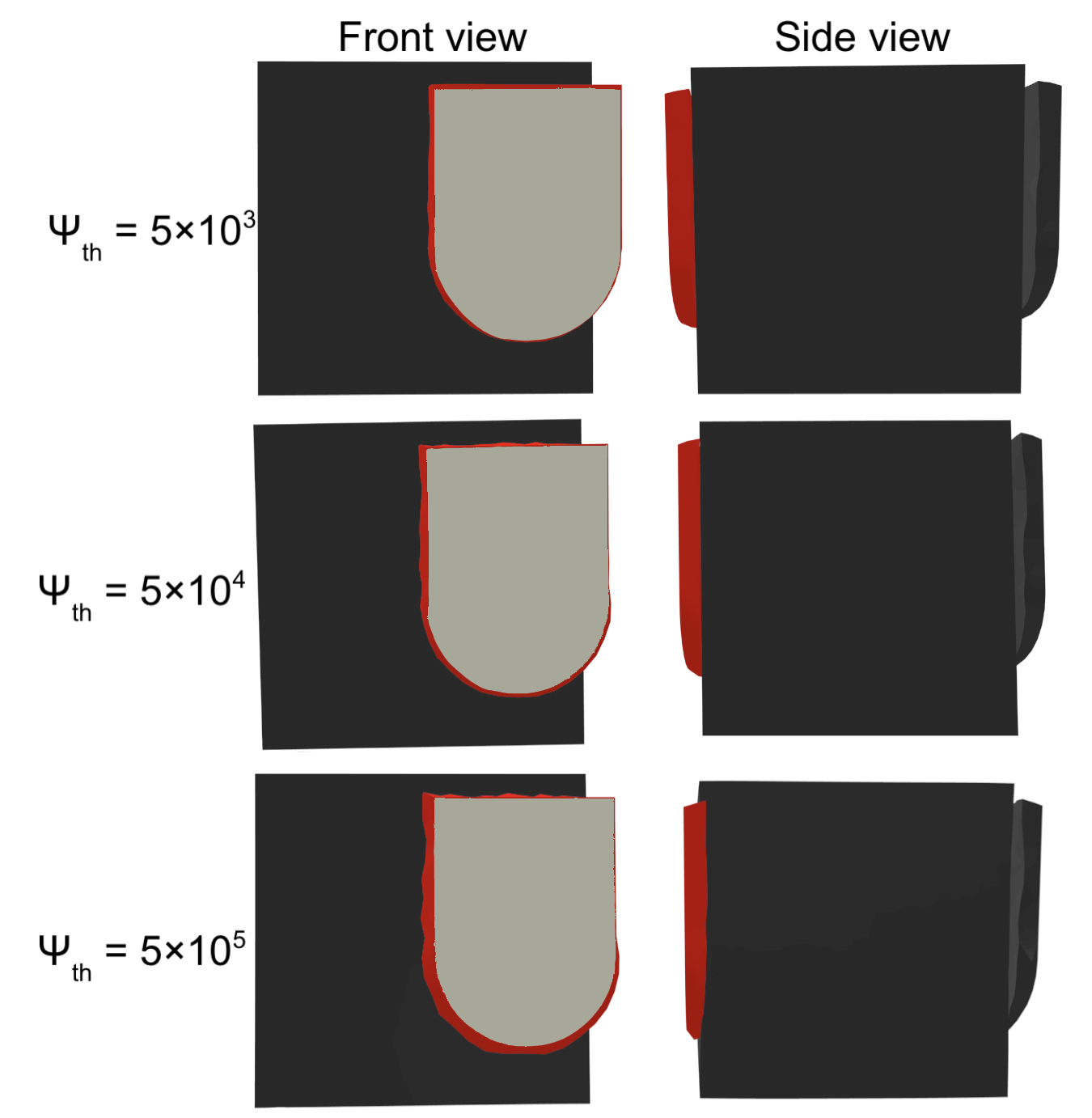}
    \caption{Tested values for $\Psi_{th}$, and the gripper pad deformation. Sides of the gripper pad are colored in red. $\Psi_{th}=5\times10^{4}$ models the gripper pad thinning under pressure, without deforming in an unrealistic way.}
    \label{fig:thres_analysis}
\end{figure}

It is critical for grasp simulators to account for uncertainty and imprecision in robot control, as there are two main sources of uncertainties when grasping in physical environments: robot position and object pose estimation. For some grasps that are very sensitive to contact area, as shown in Figure~\ref{fig:corner_grasp}, perturbations have higher chances of affecting grasp success rates. 

The calculated post-grasp object pose and grasp success in IPC-GraspSim is deterministic for each input grasp. To introduce randomness for Monte-Carlo sampling in this simulator, we inject grasp pose perturbation matrix $\varepsilon$ with normally distributed translation and rotation error. Although Isaac Gym (wiht FleX backend) is not deterministic, we inject the same perturbation effects in the simulator to mirror our changes in IPC-GraspSim. 

\begin{figure}
    \centering
    \includegraphics[width=\linewidth]{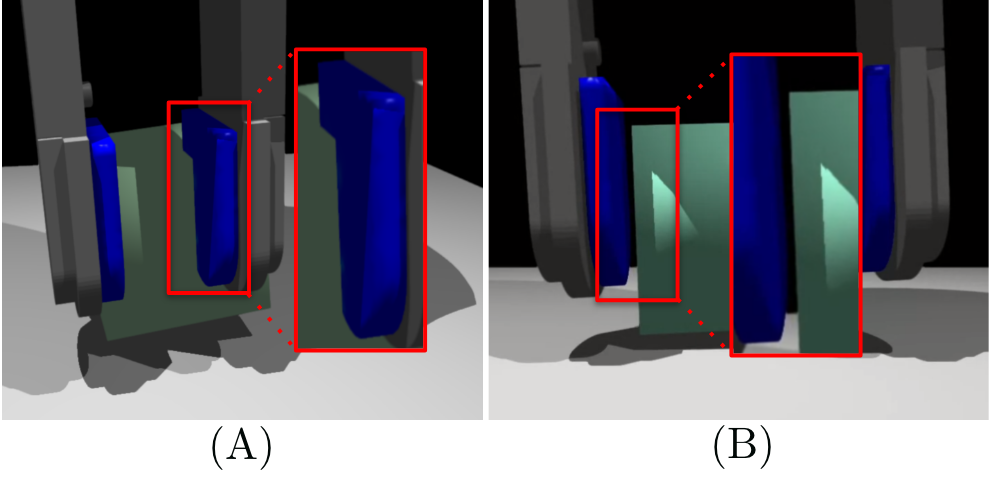}
    \caption{
    Cube grasps with incorrect contact handling: (A)~interpenetration, (B)~force-field effect. 
    }
    \label{fig:interpenetrate_issue}
\end{figure}

\subsection{Sensitivity Analysis of $\Psi_{th}$}
Due to the limitations in the IPC simulator, IPC-GraspSim uses $\Psi$, the deformation or strain energy in a given mesh, to approximate closing force applied by the compliant gripper jaws on the target object. $\Psi_{th}$ is a user-defined threshold to limit the amount of deformation in each pad ($\Psi_1$, $\Psi_2$), and corresponds to the $\Psi$ that best matches our maximum gripper closing force for the specified gripper. We attempt multiple grasps on the `cube` object with various values of $\Psi_{th}$ (shown in Figure~\ref{fig:thres_analysis}). We find that $\Psi_{th}=5\times10^{4}$ properly models the gripper pad thinning under pressure while avoiding unrealistic deformation. 

\subsection{Contact Handling in Softbody Simulators}
\begin{figure*}
    \centering
    \includegraphics[width=0.85\linewidth]{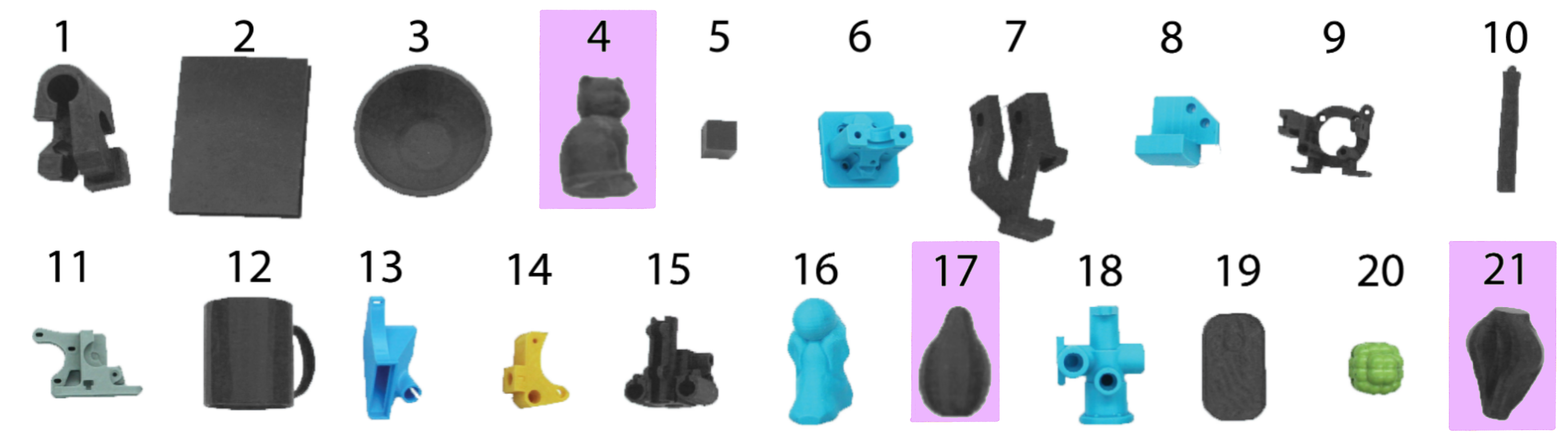}
    \caption{The 21 adversarial objects used in the grasp dataset collected by \citet{danielczuk2019reach}, where analytic grasp models perform poorly. Each object is sampled with 25 test grasps, 5 trials per grasp. The training object set (cat, pear, vase) for our experiments is highlighted in purple.
    }
    \label{fig:rep_objs}
\end{figure*}
\begin{table*}
    \centering
    \begin{tabularx}{\linewidth}{YXXXXX} \toprule
        \multirow{2}{1.2cm}{ } & \multicolumn{5}{c@{}}{\text{Cube: number of subdivisions (vertices)}} \\
    \cmidrule(l){2-6} & 0(8) & 1(26) & 2(98) & 3(399) & 4(1698) \\ \midrule
    Runtime (s/grasp) & 39 $\pm$ 27 & 40 $\pm$ 22 & 37 $\pm$ 21 & 73 $\pm$ 39 & 394 $\pm$ 287 \\ \bottomrule
    \end{tabularx}
    \caption{IPC-GraspSim runtime test with grasping simulations of cubes with varied resolutions on successful grasp in Figure 3 in the main paper, over 10 grasp trials each, on an Ubuntu 18.04 on a single core of a Dual 20-Core Intel Xeon E5-2698 v4 2.2 GHz. Number of faces per side is $2 \cdot 4^n$ for cube with $n$ subdivides.}
    \label{tab:time_results} 
\end{table*}
As mentioned in the main paper, the computationally efficient penalty-based contact methods are prone to large interpenetration effects, as shown in Figure~\ref{fig:interpenetrate_issue}A: the gripper pad intersects with the target object. These effects become more extreme when simulating softbody contacts with large Young's Modulus values $E$, or contacts between materials with large differences in $E$. 

In order to avoid large interpenetration errors, other works have modified penalty-based contact methods to calculate contact forces based on a ``thickened" version of the mesh, such that nonzero contact forces can be applied without observing any mesh penetration. However, this method may introduce ``force field" effects if the layer depth is too large: an example is shown in Figure~\ref{fig:interpenetrate_issue}B, where the gripper pad is able to push and grab onto the target object although no contact is visually observed. 

Neither of these behaviors are not acceptable in grasping contexts, as the error in Figure~\ref{fig:interpenetrate_issue}A will lead to false sticking behavior (false positive), and Figure~\ref{fig:interpenetrate_issue}B will lead to incorrect dynamics from inaccurate collision handling. 

\subsection{Grasp Dataset}
This dataset contains 21 adversarial objects where analytic grasp methods perform poorly due to large post-grasp displacements. Even for seemingly simple objects, such as object 5 (``cube") and object 10 in Figure~\ref{fig:rep_objs}, small perturbations as mentioned in Section~\ref{sec:stochastic} introduce discrepancies between pre-grasp and post-grasp contact area. However, a subset of 16 objects were chosen for our experiments due to 1) skewed success/unsuccessful grasp ratio and 2) timing constraints, as IPC-GraspSim was not optimized for speed. Out of 16, 3 objects were chosen for their 1) similar number of successful and unsuccessful grasps and 2) varied geometries. 

\subsection{IPC-GraspSim Runtime Tests} 
To study this effect in IPC-GraspSim, we choose the rigid cube, a simple object, and vary its resolution by subdividing the surface mesh before creating the volume mesh inputted to IPC. Then, we run grasping simulation on the varied meshes on a successful grasp. The runtime results are shown in Table \ref{tab:time_results}. Runtime increases almost linearly with the number of subdivisions in the cube, which is linearly proportional to the number of contacts made between the mesh surfaces. We hypothesize that the runtime is not heavily affected in the simpler models since the gripper pads are more complex than the cube themselves, and serve as the bottleneck to the code complexity. 

\section*{Acknowledgements}
\begin{footnotesize}
This research was performed at the AUTOLAB at UC Berkeley in affiliation with the Berkeley AI Research (BAIR) Lab, and the CITRIS ``People and Robots'' (CPAR) Initiative. The authors were supported in part by donations from Google, Siemens, Toyota Research Institute and Autodesk. Michael Danielczuk is supported by the National Science Foundation Graduate Research Fellowship Program under Grant No. DGE 1752814. We thank Jeffrey Ichnowski who introduced Incremental Potential Contact (IPC) to us. We also thank Minchen Li and Zachary Ferguson for support in using IPC.
\end{footnotesize}

\renewcommand*{\bibfont}{\footnotesize}
\printbibliography 
\end{document}